%% file: main.tex
\pdfoutput=1 % Keep this at the top of the file for arXiv to see
\documentclass{INTERSPEECH2023}

\interspeechcameraready 

\usepackage{tipa}
\usepackage{multirow}
\usepackage{inconsolata}
\usepackage{siunitx}
\usepackage{newtxmath}
\robustify\bfseries

\newcommand{\HPER}{PFER}

\title{Universal Automatic Phonetic Transcription \\ into the International Phonetic Alphabet}
\name{Chihiro Taguchi$^1$,
    Yusuke Sakai$^2$,
    Parisa Haghani$^3$,
    David Chiang$^1$
    }
\address{
  $^1$University of Notre Dame, United States\\
  $^2$Nara Institute of Science and Technology, Japan \\
  $^3$Google, United States}
\email{ctaguchi@nd.edu,
sakai.yusuke.sr9@is.naist.jp,
parisah@google.com,
dchiang@nd.edu
}

\begin{document}

\maketitle
 
\input{0abstract.tex}

\input{1introduction.tex}
\input{2related.tex}
\input{3method.tex}
\input{4experiments.tex}
\input{5results.tex}

\input{6discussion.tex}
\input{7conclusion.tex}

\section{Acknowledgments}
This material is based upon work supported by the National Science Foundation under Grant No.~BCS-2109709.

\goodbreak

\bibliographystyle{IEEEtran}
\bibliography{main}

\end{document}

%% file: 0abstract.tex
\begin{abstract}

This paper presents a state-of-the-art model for transcribing speech in any language into the International Phonetic Alphabet (IPA).
Transcription of spoken languages into IPA is an essential yet time-consuming process in language documentation,
and even partially automating this process has the potential to drastically speed up the documentation of endangered languages.
Like the previous best speech-to-IPA model (Wav2Vec2Phoneme), our model is based on wav2vec 2.0 and is fine-tuned to predict IPA from audio input.
We use training data from seven languages from CommonVoice 11.0, transcribed into IPA semi-automatically.
Although this training dataset is much smaller than Wav2Vec2Phoneme's, its higher quality lets our model achieve comparable or better results.
Furthermore, we show that the quality of our universal speech-to-IPA models is close to that of human annotators.
\end{abstract}

\medskip

\noindent\textbf{Index Terms}: speech recognition, phonetics, natural language processing, language documentation

%% file: 1introduction.tex
\section{Introduction}

It is estimated that there are approximately 7,000 languages in the world, at least half of which will be extinct by the next century \cite{austin2011}.
To record and revitalize these endangered languages, there have been many efforts by field linguists to document them and provide grammatical sketches and dictionaries.
However, manual transcription of the recorded audio consumes a large amount of time in the documentation process \cite{thieberger2017}.
Given this situation, the technology of automatic speech recognition (ASR) has the potential to accelerate language documentation.

This paper presents a model that converts speech to the International Phonetic Alphabet (IPA), which is the standard used in phonemic and phonetic transcription of understudied languages.
We introduce a simple yet high-performance speech-to-IPA model that outperforms the current state-of-the-art, Wav2Vec2Phoneme \cite{xu-2022}.
Like Wav2Vec2Phoneme, our model is based on wav2vec 2.0 \cite{baevski2020} and is fine-tuned to predict IPA from audio input.
We use training data from seven languages in the CommonVoice\footnote{\url{https://commonvoice.mozilla.org}} dataset, transcribed into IPA semi-automatically; that is, the quality of IPA transcription tools is verified to be reliable enough.
The experimental results show that our model with 1,000 training samples per language (about 9 hours in total) outperforms existing speech-to-IPA models trained on more samples with more languages.
The contributions of this paper are summarized as follows:
\begin{itemize}
    \item We developed a new state-of-the-art speech-to-IPA model with a small training dataset;
    \item The results suggest that the quality of phonetic IPA labels and the linguistic diversity in the training dataset are significant factors of the performance in this task;
    \item The models, datasets, and G2P tools developed for this study will be publicly available.\footnote{\url{https://github.com/ctaguchi/multipa}}
\end{itemize}

%% file: 2related.tex
\section{Related Work} 
Multilingual ASR has been advancing towards the ultimate goal of universal transcription for all spoken languages~\cite{conneau2021, li2022, whisper, zhang2023}.
However, mainstream objectives of ASR models have been to transcribe speech into graphemes, which are dependent on each language and its orthography; in other words, they are largely language-dependent rather than language-universal.
Since endangered languages tend not to have any established orthography, they are often marginalized from computational processing methods that are based on orthography.

Some efforts have been made to train multilingual ASR models to transcribe low-resource languages into phones or phonemes written in IPA \cite{schultz1997, cohen1997, schultz2001, dalmia2018}.
In particular, Michaud \cite{michaud2018} has carried out a successful case study of applying ASR to the documentation of the Na language.
In recent years, more work on multilingual speech-to-phone ASR has been investigated using neural network models.
Allosaurus \cite{li2020universal} is a speech-to-IPA tool that is aimed to transcribe more than 2,000 languages.
Gao et al. \cite{gao2021} train a wav2vec-based model on about 1,500 hours of audio from 15 languages; the text transcriptions are converted to IPA using a grapheme-to-phoneme (G2P) system.
Wav2Vec2Phoneme \cite{xu-2022} is a wav2vec2-based model that has a similar approach to ours.
It fine-tunes wav2vec2-large-xlsr-53~\cite{conneau2021} on approximately 2,000 hours of audio from 26 languages from CommonVoice.
A significant difference between our approach and these models is the preprocessing phase to generate labels in IPA.
Their IPA transcription labels are fully automatically generated using rule-based G2P tools, which do not necessarily guarantee the quality of its phonetic or phonemic transcription, while we only used G2P tools that are verified to be reliable enough.
The details of the preprocessing step are described in the next section.

%% file: 3method.tex
\section{Method}
\subsection{Data}
\label{sec:data}

Datasets containing audio with close transcriptions into IPA are extremely scarce.
Linguistic resources with speech--IPA pairs such as Pangloss \cite{michailovsky2014} often contain IPA transcriptions that are not phonetic but phonemic or use language-specific conventions.
To overcome this difficulty, most existing speech-to-IPA models \cite{xu-2022, li2020universal, gao2021}
use automatic G2P tools to transliterate the text of the datasets into IPA\@.
However, because the output of these G2P tools is phonemic, and their quality is not guaranteed, it is possible that the generated IPA transcriptions do not represent the actual pronunciation accurately.

In this study, to alleviate this problem, we semi-automated the IPA transcription in two ways.
First, we manually checked the quality of the G2P tools and only used those that were reliable enough.
Second, to leverage the utility of the G2P tools as much as possible, we particularly chose languages that have 
a consistent orthography-to-pronunciation mapping.
Based on these policies, we picked seven languages (Japanese, Polish, Maltese, Hungarian, Finnish, Greek, and Tamil) from CommonVoice 11.0.\footnote{\url{https://huggingface.co/datasets/mozilla-foundation/common_voice_11_0}}
For G2P, we used a combination of Epitran\footnote{\url{https://github.com/dmort27/epitran}}~\cite{mortensen2018} and tools implemented by us.
Non-phonetic characters including punctuation symbols and spaces were removed at this point.
Also, tones and other suprasegmental elements were not included in the transcription, because describing all the pitch information for each syllable of any languages would have yielded too much unnecessary information for further linguistic analysis.
Table \ref{tab:datasets} summarizes the datasets used for training and validation.

All audio clips were downsampled to 16 kHz for training.
Audio clips longer than 6 seconds were removed from the training and validation sets because they can exhaust available memory.
We also removed audio samples of low quality that are labeled with more than one negative vote.
After this preprocessing, we prepared three sets of training and validation data with different sample sizes.
The first set contains 1k training samples and 200 validation samples per language, the second set 2k training samples and 400 validation samples per language, and the third set all the training samples available in the dataset.
The samples were randomly selected from the preprocessed datasets, following the dataset splits provided by CommonVoice.
For the evaluation of the model performance in supervised settings, i.e., where the languages used in the test are learned during the training,
we picked 100 samples per language (i.e., 700 samples in total) from the test split.

\begin{table}[t]
    \caption{A description of the datasets used for the training and validation sets in the study.
    ``\# Train'' in the middle column refers to the maximum size of the training set available in CommonVoice 11.0.
    Seven languages are picked from CommonVoice for training the model.}
    \label{tab:datasets}
    \centering
    \begin{tabular}{lrl} \toprule
        CommonVoice & \# Train & G2P tool \\ \midrule
        Japanese (ja) & 6,150 & Manual rules \\
        Polish (pl) & 15,419 & Epitran \\
        Maltese (mt) & 1,780 & Manual rules \\
        Hungarian (hu) & 7,044 & Manual rules \\
        Finnish (fi) & 2,044 & Manual rules \\
        Greek (el) & 1,722 & Manual rules \\
        Tamil (ta) & 16,189 & Epitran \\ \midrule
        Total & 50,348 & \\
        \bottomrule
    \end{tabular}
\end{table}

For testing the universality of the models, in addition to the in-domain evaluation of the performance in the trained languages, we picked four low-resource languages from CommonVoice that are typologically diverse, geographically distant, and genetically unrelated: Luganda ($<$ Bantu; Uganda), Upper Sorbian ($<$ Indo-European; Saxony, Germany), Hakha Chin~($<$ Sino-Tibetan; Chin State, Myanmar), and Tatar ($<$ Turkic; Tatarstan, Russia).
For annotation, we hired two language technology graduate students of their 20s who had had training in phonetics and IPA transcription, and we had them transcribe the test audio clips manually as mock fieldworkers.
Their first language was Japanese.  
To prevent the annotator from guessing a language, they were not told what and how many languages were included in the samples.
These clips were randomly chosen from these four languages, and the annotator transcribed the assigned audio clips until we had 100 annotated samples.
We used the transcriptions by one of the two annotators as the gold labels in the evaluation, and those by the other were used to measure the inter-annotator agreement (IAA) between the two annotators with the metrics used in this study (see Section~\ref{sec:metrics}).

\begin{table}[t]
    \caption{A description of the test set for testing the model on unseen languages.
    We picked four typologically diverse low-resource languages that are not part of the training dataset.
    A trained human annotator transcribed randomly chosen audio samples up to 100 samples in total.}
    \label{tab:datasets-test}
    \centering
    \begin{tabular}{lrl} \toprule
        CommonVoice & \# Samples & G2P tool \\ \midrule
        Luganda (lg) & 22 & Manual annotation \\
        Upper Sorbian (hsb) & 24 & Manual annotation \\
        Hakha Chin (cnh) & 25 & Manual annotation \\
        Tatar (tt) & 29 & Manual annotation \\ 
        \bottomrule
    \end{tabular}
\end{table}

\subsection{Model}
The pretrained model is wav2vec2-large-xlsr-53~\cite{conneau2021} which was trained on 56k hours of speech data with 53 languages from CommonVoice, Multilingual LibriSpeech, and BABEL\@. 
Our objective is to fine-tune the model to predict the IPA string of audio input.
We implement the fine-tuned model with \texttt{Wav2Vec2ForCTC} provided in the \texttt{transformers} library to train it with the Connectionist Temporal Classification (CTC) loss~\cite{graves2006}.
Since our goal is to train a model applicable to unseen languages, we did not use encoder-decoder models that are not language-agnostic such as Whisper~\cite{whisper}.

%% file: 4experiments.tex
\section{Experiments}
% Results
\input{in-domain.tex}
\input{zero-shot.tex}

\subsection{Setup}
We used \texttt{Wav2Vec2CTCTokenizer} as the tokenizer and included in the vocabulary the full list of the IPA characters and their possible combinations such as multi-letter phones and co-articulated consonants.
The full IPA list, obtained from PanPhon \cite{mortensen2016}, consists of 6,487 phones.

We set the CTC loss reduction (\texttt{ctc\_loss\_reduction}) to \texttt{mean}, the learning rate to \num{3e-4}, the warmup steps to 500, the number of training epochs to 30; other numerical hyperparameters used the default values defined in the configuration in the \texttt{transformers} library as of version 4.26.0.
We used 150 training epochs for the extremely low-resource setting with 100 samples per language and 5 epochs for the setting with full training samples, so that the models would learn to output IPA sequences well enough while keeping them from overfitting.
The feature extractor of the pretrained model is frozen before fine-tuning.

In the default setting (1k training samples per language), the training took $\sim$4 hours in runtime with four GTX1080Ti GPUs, and the average power consumption was $\sim$7.5kW.

\subsection{Evaluation metrics}
\label{sec:metrics}

We compare our models with the two existing speech-to-IPA models: Allosaurus \cite{li2020universal} and Wav2Vec2Phoneme \cite{xu-2022}.
We evaluate these models with two metrics: naïve phone error rate~(PER) and PER with phonetic feature edit distance \cite{mortensen2016}, which we call \HPER{} (phone feature error rate) for short.
PER is a modification of character error rate (CER) whose basic unit is a phone (which may consist of more than one Unicode character) instead of a character.
However, a problem of using PER in this task is that PER ignores phonetic similarities among phones.
For example, suppose that a model predicts {[}k\super h\ae t{]} for a spoken word with the label {[}k\ae t{]}, where they express the first consonant ({[}k{]} or~{[}k\super h{]}) in a different manner.
They only differ in one phonetic feature, namely aspiration, and share a similar acoustic impression;
therefore, intuitively speaking, their distance should be smaller than that of two utterly unrelated phones like [k] and~[a].
PER treats both pairs of phones equally.
\HPER, in contrast, considers acoustic similarities among phones by representing each phone as a collection of phonetic features.
In our experiments, we used the implementation provided in PanPhon \cite{mortensen2016}, which defines 24 phonetic features for each phone.
\HPER{} calculates the Hamming distance of features between two phones so that one feature mismatch has a cost of 1/24.\footnote{The documentation of PanPhon at the time of writing notes that the cost of a feature mismatch is 1/22, but the implementation gives 1/24 because their feature table contains 24 features.}
Other string operations (insertion and substitution) cost 1.
For this reason, we put more emphasis on the \HPER-based comparison than PER in this study because \HPER{} is more representative of the transcription accuracy.

%% file: in-domain.tex
\begin{table*}[t]
    \caption{The evaluation scores in the supervised setting for models with 1k, 2k, and full training samples.
    }
    \label{tab:in-domain}
    \centering
    \begin{tabular}{llSSSSSSSS}\toprule
        Metric & \# Train samples & {Japanese} & {Polish} & {Maltese} & {Hungarian} & {Finnish} & {Greek} & {Tamil} & {Overall} \\ \midrule
        \multirow{3}{4.3em}{PER (\%)} & 1k & 17.363 & 16.504 & 14.337 & 44.902 & 29.615 & 24.734 & 26.855 & 24.901 \\
        & 2k & 12.799 & 12.324 & 10.370 & 43.068 & 28.343 & 23.729 & 25.823 & 22.351 \\ 
        & full & 47.779 & 7.067 & 14.939 & 42.285 & 13.755 & 7.662 & 13.441 & 20.990 \\ \midrule
        \multirow{3}{4.3em}{\HPER{} (\%)}  & 1k & 4.813 & 4.500 & 5.162 & 11.663 & 8.195 & 8.449 & 6.185 & 6.995 \\
        & 2k & \bfseries 3.635 & 3.841 & \bfseries 3.479 & \bfseries 10.830 & 8.037 & 7.683 & 5.704 & 6.173 \\
        & full & 8.88 & \bfseries 2.522 & 4.876 & \bfseries 10.788 & \bfseries 4.978 & \bfseries 4.098 & \bfseries 3.649 & \bfseries 5.684 \\
        \bottomrule
    \end{tabular}
\end{table*}

%% file: zero-shot.tex
\begin{table*}[t]
    \caption{Comparison of the scores in the zero-shot setting.
    The numbers in parentheses next to our models refer to the number of training samples per language.
    Human (IAA) is the IAA scores between the two human annotators.}
    \label{tab:zero-shot}
    \centering
    \begin{tabular}{llSSSSS} \toprule
        Metric & Model & {Luganda} & {Upper Sorbian} & {Hakha Chin} & {Tatar} & {Overall} \\ \midrule
        \multirow{6}{4.3em}{PER (\%)} & Allosaurus & 104.096 & 93.917 & 79.375 & 89.648 & 91.759 \\
        & Wav2Vec2Phoneme & 64.044 & 66.105 & 70.046 & 62.998 & 65.798 \\
        & Ours (1k) & 74.013 & 68.608 & 72.976 & 67.710 & 70.827 \\ 
        & Ours (2k) & 76.978 & 69.430 & 72.699 & 67.454 & 71.640 \\
        & Ours (full) & 70.891 & 69.792 & 68.691 & 63.182 & 63.182 \\
        & Human (IAA) & 52.915 & 52.486 & 55.253 & 52.676 & 53.333 \\ \midrule
        \multirow{6}{4.3em}{\HPER{} (\%)} & Allosaurus & 46.073 & 36.297 & 36.297 & 30.050 & 34.247 \\
        & Wav2Vec2Phoneme & 24.186 & 26.055 & \bfseries 19.257 & 20.013 & 22.378 \\
        & Ours (1k) & \bfseries 20.790 & 23.977 & 21.292 & \bfseries 18.825 & \bfseries 21.221 \\
        & Ours (2k) & 22.744 & 24.922 & 21.801 & 19.389 & 22.214 \\
        & Ours (full) & 23.032 & \bfseries 23.105 & 20.320 & \bfseries 18.760 & 21.304 \\
        & Human (IAA) & 19.257 & 22.129 & 17.765 & 19.132 & 19.571 \\
        \bottomrule
    \end{tabular}
\end{table*}

%% file: 5results.tex
\section{Results}

This section reports the results of the performance of our models and compares them to the two existing speech-to-IPA models, Allosaurus and Wav2Vec2Phoneme, as well as the human IAA\@.

\subsection{In-domain evaluation}
Table \ref{tab:in-domain} reports the performance of automatic IPA transcription for the languages the models were trained with.
In both PER and \HPER, our model performed better when the training size is larger.
In the overall scores, the model with the maximum training size (50,348 samples, or $\sim$60 hours, in total) achieved \num{5.68}\% \HPER, while the models with 1k and 2k training samples per language scored \num{7.00}\% and \num{6.17}\% \HPER, respectively.
These results follow the general assumption in machine learning that training with more samples gives better performance.

\subsection{Zero-shot evaluation}
\label{sec:zero-shot}
Table \ref{tab:zero-shot} compares the scores of the performance in transcribing languages unseen during the training phase.
Overall, our model with 1k training samples per language (\num{21.221}\% \HPER) performed the best among the tested models.
In particular, it outperformed both of Allosaurus (\num{34.25}\% \HPER) and Wav2Vec2Phoneme (\num{22.38}\% \HPER).
Interestingly, our model with 1k training samples per language performed better on average than that with 2k samples and full samples despite its smaller size.
Given that the overall human IAA rate is \num{19.571}\% \HPER, our models and Wav2Vec2Phoneme have nearly reached performance comparable with human annotators.
For reference, Table \ref{tab:example} compares the actual output by Allosaurus, Wav2Vec2Phoneme, and our model (1k).

\begin{table}[t]
    \centering
    \caption{Output by Allosaurus, Wav2Vec2Phoneme, and our model (1k).
    The sample is extracted from the Luganda validation set in CommonVoice.}
    \begin{tabular}{@{}ll@{}} \toprule
        Sentence & Omukama mulungi obudde bwonna. \\ \midrule
        Allosaurus & omu\textschwa k\super hamamu\textschwa\textfishhookr on\texttoptiebar{d\textyogh}io\textinvscr u\textschwa d\super je\textlowering{b}u\textschwa oen\textlengthmark \textturnv \\
        Wav2Vec2Phoneme & omukamamudund\textyogh ipovudevonna \\
        Ours (1k) & omukamamu\textfishhookr un\texttoptiebar{d\textyogh}ipoputd\textepsilon vonna \\
        \bottomrule
    \end{tabular}
    \label{tab:example}
\end{table}

%% file: 6discussion.tex
\section{Discussion}
The above results show that the performance of Wav2Vec2Phoneme and our models is comparable to human annotators, while still lagging behind by approximately~2 to 4\% in \HPER\@.
The better performance of our model (1k) than Wav2Vec2Phoneme is particularly remarkable because ours was only trained on 7k samples ($\sim$9.7 hours) from seven languages, while Wav2Vec2Phoneme uses at least 100 hours of audio from at least 26 languages from CommonVoice.
In addition, these results suggest that transfer learning for the speech-to-IPA task reaches a performance plateau with a rather small training dataset, and increasing the number of samples would not contribute to improvement.
In the following ablation studies, we discuss what are the factors that affect performance.

\subsection{Ablation studies}
We considered four parameters in this ablation: training samples per language (Size), additional data from Forvo\footnote{\url{https://forvo.com}}~(+/$-$Forvo), quality filtering (+/$-$QF), and linguistic diversity in the training data (\# Languages).
We used the same evaluation methods as in the zero-shot evaluation in Section~\ref{sec:zero-shot}.
We prepared an additional small dataset with 353 samples from six languages (Adyghe, Arabic, Burmese, Icelandic, Xhosa, and Zulu) retrieved from Forvo to add phonetic and linguistic diversity to the training data.
However, a numerical comparison in Table \ref{tab:ablation} shows no benefit from adding the Forvo data.
It may be that the additional data size is too small to affect the performance, and further investigation is called for.

Within the setting with filtering out poor-quality audio~(+QF in Table \ref{tab:ablation}), the extremely low-resource setting with only 100 samples per language (i.e., 700 samples in total) performed the worst; however, the results with 1k samples were better than that with 2k samples.
This implies that increasing the amount of training data does not promise a performance improvement and that it might hit a plateau before 1k samples per language.
Table \ref{tab:ablation} also shows that removing audio files of poor quality gave us better results with 1k samples per language but did worse with 2k samples per language. This suggests that removing audio files of poor quality does not necessarily promise an improvement, either.

Table \ref{tab:ablation-lang} specifically compares the effect of linguistic diversity in the training data.
We can see that having more diversity provides better results.
This is expected since having fewer languages in the training set means that predictions will be strictly limited to the phonetic inventory of those languages.

\input{ablation.tex}

\input{ablation-lang.tex}

\subsection{Limitations}
Last but not least, we mention several limitations that this study has faced but can be improved in future research.
First, we could only use the seven languages that can be automatically transcribed from graphemes to phones relatively easily and accurately (Section \ref{sec:data}).
The languages used to train the model are still too few to reflect the actual linguistic diversity of the world.
Second, it is not a perfect solution to rely on rule-based G2P tools to generate IPA transcriptions to be used as the labels, because our objective is to generate IPA transcription as accurately as human transcription.
Third, due to the difficulty of hiring well-trained annotators for IPA transcription, our test data for the zero-shot evaluation only contains~100 samples in total.
We can overcome these three limitations by developing larger high-quality datasets with speech and phonetic transcriptions.
Fourth, our models do not consider tones as phonetic features.
Extensions to include lexical and grammatical tones can be done by fine-tuning the model to the phonology of a specific tonal language; or, more generally, it might be possible to incorporate tones in the system by adding a layer or token that implicitly identifies whether the language is tonal or not.

%% file: ablation.tex
\begin{table}[t]
    \caption{
    Ablation studies on quality filtering and additional data.
    The unit is PFER (\%).
    ``+/$-$Forvo'' means whether the Forvo dataset was included in training, and ``+/$-$QF'' whether low-quality audio samples were removed.
    The Forvo dataset was not applied to the low-resource setting (Size$=$100).}
    \label{tab:ablation}
    \centering
    \begin{tabular}{lSSSS} \toprule
         & \multicolumn{2}{c}{+Forvo} & \multicolumn{2}{c}{$-$Forvo} \\
        Size & {+QF} & {$-$QF} & {+QF} & {$-$QF} \\ \midrule
        100 & {---} & {---} & 24.257 & 22.624 \\
        1k & 22.618 & 21.430 & 21.221 & 21.220 \\
        2k & 21.767 & 22.537 & 22.214 & 22.873 \\
        \bottomrule
    \end{tabular}
\end{table}

%% file: ablation-lang.tex
\begin{table}[t]
    \caption{Ablation studies on the effect of  linguistic diversity in the training data.
    The unit is PFER (\%).
    All models were trained without audio of bad quality and the Forvo dataset.
    }
    \label{tab:ablation-lang}
    \centering
    \begin{tabular}{lS} \toprule
        \# Languages & {\HPER{} (\%)} \\ \midrule
        1 (ja) & 28.3 \\
        3 (ja, pl, mt) & 24.1 \\
        7 (all) & 21.2 \\
        \bottomrule
    \end{tabular}
\end{table}

%% file: 7conclusion.tex
\section{Conclusion}
This study introduced a new universal speech-to-IPA model trained with only 7k samples from seven languages.
We showed that our model with 1k training samples per language performs better than existing speech-to-IPA models trained on larger datasets from at least 26 languages.
Our model achieved \num{21.2}\% in \HPER, which was almost comparable to the human IAA score.
In our settings, it was observed that broader linguistic diversity in the training data gives more accurate IPA transcription.
Our results also suggested the importance of using clean phonetic transcription in the training dataset.

%% file: main.bbl
% Generated by IEEEtran.bst, version: 1.13 (2008/09/30)
\begin{thebibliography}{10}
\providecommand{\url}[1]{#1}
\csname url@samestyle\endcsname
\providecommand{\newblock}{\relax}
\providecommand{\bibinfo}[2]{#2}
\providecommand{\BIBentrySTDinterwordspacing}{\spaceskip=0pt\relax}
\providecommand{\BIBentryALTinterwordstretchfactor}{4}
\providecommand{\BIBentryALTinterwordspacing}{\spaceskip=\fontdimen2\font plus
\BIBentryALTinterwordstretchfactor\fontdimen3\font minus
  \fontdimen4\font\relax}
\providecommand{\BIBforeignlanguage}[2]{{%
\expandafter\ifx\csname l@#1\endcsname\relax
\typeout{** WARNING: IEEEtran.bst: No hyphenation pattern has been}%
\typeout{** loaded for the language `#1'. Using the pattern for}%
\typeout{** the default language instead.}%
\else
\language=\csname l@#1\endcsname
\fi
#2}}
\providecommand{\BIBdecl}{\relax}
\BIBdecl

\bibitem{austin2011}
\BIBentryALTinterwordspacing
P.~Austin and J.~Sallabank, Eds., \emph{The Cambridge Handbook of Endangered
  Languages}, ser. Cambridge Handbooks in Language and Linguistics.\hskip 1em
  plus 0.5em minus 0.4em\relax Cambridge University Press, 2011. [Online].
  Available: \url{https://doi.org/10.1017/CBO9780511975981}
\BIBentrySTDinterwordspacing

\bibitem{thieberger2017}
N.~Thieberger, ``{LD\&C} possibilities for the next decade,'' \emph{Language
  Documentation \& Conservation}, vol.~11, pp. 1--4, 2017.

\bibitem{xu-2022}
\BIBentryALTinterwordspacing
Q.~Xu, A.~Baevski, and M.~Auli, ``Simple and effective zero-shot cross-lingual
  phoneme recognition,'' in \emph{Proceedings of Interspeech}, 2022, pp.
  2113--2117. [Online]. Available:
  \url{https://doi.org/10.21437/Interspeech.2022-60}
\BIBentrySTDinterwordspacing

\bibitem{baevski2020}
\BIBentryALTinterwordspacing
A.~Baevski, H.~Zhou, A.~Mohamed, and M.~Auli, ``{wav2vec} 2.0: A framework for
  self-supervised learning of speech representations,'' in \emph{Proceedings of
  the 34th International Conference on Neural Information Processing Systems},
  2020, pp. 12\,449--12\,460. [Online]. Available:
  \url{https://arxiv.org/abs/2006.11477}
\BIBentrySTDinterwordspacing

\bibitem{conneau2021}
\BIBentryALTinterwordspacing
A.~Conneau, A.~Baevski, R.~Collobert, A.~Mohamed, and M.~Auli, ``Unsupervised
  cross-lingual representation learning for speech recognition,'' in
  \emph{Proceedings of Interspeech}, 2021, pp. 2426--2430. [Online]. Available:
  \url{https://doi.org/10.21437/Interspeech.2021-329}
\BIBentrySTDinterwordspacing

\bibitem{li2022}
\BIBentryALTinterwordspacing
B.~Li, R.~Pang, Y.~Zhang, T.~N. Sainath, T.~Strohman, P.~Haghani, Y.~Zhu,
  B.~Farris, N.~Gaur, and M.~Prasad, ``Massively multilingual {ASR}: A lifelong
  learning solution,'' in \emph{Proceedings of the 2022 IEEE International
  Conference on Acoustics, Speech and Signal Processing (ICASSP)}, 2022, pp.
  6397--6401. [Online]. Available:
  \url{https://doi.org/10.1109/ICASSP43922.2022.9746594}
\BIBentrySTDinterwordspacing

\bibitem{whisper}
\BIBentryALTinterwordspacing
A.~Radford, J.~W. Kim, T.~Xu, G.~Brockman, C.~McLeavey, and I.~Sutskever,
  ``Robust speech recognition via large-scale weak supervision,'' 2022.
  [Online]. Available: \url{https://arxiv.org/abs/2212.04356}
\BIBentrySTDinterwordspacing

\bibitem{zhang2023}
\BIBentryALTinterwordspacing
Y.~Zhang, W.~Han, J.~Qin, Y.~Wang, A.~Bapna, Z.~Chen, N.~Chen, B.~Li,
  V.~Axelrod, G.~Wang, Z.~Meng, K.~Hu, A.~Rosenberg, R.~Prabhavalkar, D.~S.
  Park, P.~Haghani, J.~Riesa, G.~Perng, H.~Soltau, T.~Strohman, B.~Ramabhadran,
  T.~Sainath, P.~Moreno, C.-C. Chiu, J.~Schalkwyk, F.~Beaufays, and Y.~Wu,
  ``{G}oogle {USM}: Scaling automatic speech recognition beyond 100
  languages,'' 2023. [Online]. Available:
  \url{https://arxiv.org/abs/2303.01037}
\BIBentrySTDinterwordspacing

\bibitem{schultz1997}
\BIBentryALTinterwordspacing
T.~Schultz and A.~Waibel, ``Fast bootstrapping of lvcsr systems with
  multilingual phoneme sets,'' in \emph{Proceedings of 5th European Conference
  on Speech Communication and Technology (EUROSPEECH '97)}, vol.~1, September
  1997, pp. 371 -- 373. [Online]. Available:
  \url{https://www.isca-speech.org/archive_v0/eurospeech_1997/e97_0371.html}
\BIBentrySTDinterwordspacing

\bibitem{cohen1997}
\BIBentryALTinterwordspacing
P.~Cohen, S.~Dharanipragada, J.~Gros, M.~Monkowski, C.~Neti, S.~Roukos, and
  T.~Ward, ``Towards a universal speech recognizer for multiple languages,'' in
  \emph{Proceedings of the 1997 IEEE Workshop on Automatic Speech Recognition
  and Understanding (ASRU)}, 1997, pp. 591--598. [Online]. Available:
  \url{https://doi.org/10.1109/ASRU.1997.659140}
\BIBentrySTDinterwordspacing

\bibitem{schultz2001}
\BIBentryALTinterwordspacing
T.~Schultz and A.~Waibel, ``Language-independent and language-adaptive acoustic
  modeling for speech recognition,'' \emph{Speech Communication}, vol.~35,
  no.~1, pp. 31--51, 2001. [Online]. Available:
  \url{https://www.sciencedirect.com/science/article/pii/S0167639300000947}
\BIBentrySTDinterwordspacing

\bibitem{dalmia2018}
\BIBentryALTinterwordspacing
S.~Dalmia, R.~Sanabria, F.~Metze, and A.~W. Black, ``Sequence-based
  multi-lingual low resource speech recognition,'' in \emph{2018 IEEE
  International Conference on Acoustics, Speech and Signal Processing
  (ICASSP)}.\hskip 1em plus 0.5em minus 0.4em\relax IEEE Press, 2018, p.
  4909–4913. [Online]. Available:
  \url{https://doi.org/10.1109/ICASSP.2018.8461802}
\BIBentrySTDinterwordspacing

\bibitem{michaud2018}
\BIBentryALTinterwordspacing
A.~Michaud, O.~Adams, T.~A. Cohn, G.~Neubig, and S.~Guillaume, ``Integrating
  automatic transcription into the language documentation workflow: Experiments
  with {N}a data and the {P}ersephone toolkit,'' \emph{Language Documentation
  \& Conservation}, vol.~12, pp. 481--513, 2018. [Online]. Available:
  \url{http://hdl.handle.net/10125/24793}
\BIBentrySTDinterwordspacing

\bibitem{li2020universal}
\BIBentryALTinterwordspacing
X.~Li, S.~Dalmia, J.~Li, M.~Lee, P.~Littell, J.~Yao, A.~Anastasopoulos, D.~R.
  Mortensen, G.~Neubig, A.~W. Black, and M.~Florian, ``Universal phone
  recognition with a multilingual allophone system,'' in \emph{Proceedings of
  the IEEE International Conference on Acoustics, Speech and Signal Processing
  (ICASSP)}, 2020, pp. 8249--8253. [Online]. Available:
  \url{https://doi.org/10.1109/ICASSP40776.2020.9054362}
\BIBentrySTDinterwordspacing

\bibitem{gao2021}
\BIBentryALTinterwordspacing
H.~Gao, J.~Ni, Y.~Zhang, K.~Qian, S.~Chang, and M.~Hasegawa-Johnson,
  ``Zero-shot cross-lingual phonetic recognition with external language
  embedding,'' in \emph{Proceedings of Interspeech}, 2021, pp. 1304--1308.
  [Online]. Available: \url{https://doi.org/10.21437/Interspeech.2021-1843}
\BIBentrySTDinterwordspacing

\bibitem{michailovsky2014}
\BIBentryALTinterwordspacing
B.~Michailovsky, M.~Mazaudon, A.~Michaud, S.~Guillaume, A.~François, and
  E.~Adamou, ``Documenting and researching endangered languages: The {P}angloss
  {C}ollection,'' \emph{Language Documentation \& Conservation}, vol.~8, pp.
  119--135, 2014. [Online]. Available: \url{http://hdl.handle.net/10125/4621}
\BIBentrySTDinterwordspacing

\bibitem{mortensen2018}
\BIBentryALTinterwordspacing
D.~R. Mortensen, S.~Dalmia, and P.~Littell,
  ``\BIBforeignlanguage{english}{{E}pitran: Precision {G2P} for many
  languages},'' in \emph{\BIBforeignlanguage{english}{Proceedings of the
  Eleventh International Conference on Language Resources and Evaluation (LREC
  2018)}}, 2018. [Online]. Available: \url{https://aclanthology.org/C16-1328}
\BIBentrySTDinterwordspacing

\bibitem{graves2006}
\BIBentryALTinterwordspacing
A.~Graves, S.~Fern\'{a}ndez, F.~Gomez, and J.~Schmidhuber, ``Connectionist
  temporal classification: Labelling unsegmented sequence data with recurrent
  neural networks,'' in \emph{Proceedings of the 23rd International Conference
  on Machine Learning}, 2006, p. 369–376. [Online]. Available:
  \url{https://doi.org/10.1145/1143844.1143891}
\BIBentrySTDinterwordspacing

\bibitem{mortensen2016}
\BIBentryALTinterwordspacing
D.~R. Mortensen, P.~Littell, A.~Bharadwaj, K.~Goyal, C.~Dyer, and L.~S. Levin,
  ``{P}an{P}hon: {A} resource for mapping {IPA} segments to articulatory
  feature vectors,'' in \emph{Proceedings of {COLING} 2016, the 26th
  International Conference on Computational Linguistics: Technical Papers},
  2016, pp. 3475--3484. [Online]. Available:
  \url{https://aclanthology.org/C16-1328}
\BIBentrySTDinterwordspacing

\end{thebibliography}
